\documentclass[lettersize,journal]{IEEEtran}
\usepackage{amsmath,amsfonts}
\usepackage{array}
\usepackage[caption=false,font=normalsize,labelfont=sf,textfont=sf]{subfig}
\usepackage{textcomp}
\usepackage{stfloats}
\usepackage{url}
\usepackage{verbatim,enumitem}

\usepackage{amsmath,amsfonts,bm}

\def\1{\bm{1}}

\DeclareMathAlphabet{\mathsfit}{\encodingdefault}{\sfdefault}{m}{sl}
\SetMathAlphabet{\mathsfit}{bold}{\encodingdefault}{\sfdefault}{bx}{n}

\usepackage{algorithm}
\usepackage{algpseudocode}
\usepackage{hyperref}
\usepackage{url}
\usepackage{makecell}
\usepackage{multirow}
\usepackage{hhline}
\usepackage{colortbl}
\usepackage{adjustbox}
\usepackage{hyperref}       
\usepackage{booktabs}       
\usepackage{nicefrac}       
\usepackage{microtype}      
\usepackage{xcolor}         
\usepackage{amsthm}
\newcommand{\cov}{\mathbb{C}ov}
\newcommand{\var}{\mathbb{V}ar}
\newcommand{\shortname}{\texttt{FedCVR-Bolt}} 
\newtheorem{proposition}{Proposition}
\newtheorem{lemma}{Lemma}

\newtheorem{remark}{Remark}
\newtheorem{theorem}{Theorem}
\newtheorem{assumption}{Assumption}
\usepackage{graphicx}
\def\BibTeX{{\rm B\kern-.05em{\sc i\kern-.025em b}\kern-.08em
    T\kern-.1667em\lower.7ex\hbox{E}\kern-.125emX}}
\usepackage{balance}
\begin{document}
\title{Tackling Heterogeneity in Federated Learning via Variance-Reduced Boltzmann Sampling within Homogeneous Social Coalitions}
\author{
Alessandro Licciardi,~\IEEEmembership{Member,~IEEE,} Roberta Raineri,~\IEEEmembership{Member,~IEEE,} Anton Proskurnikov,~\IEEEmembership{Senior Member,~IEEE,} Lamberto Rondoni,
Lorenzo Zino,~\IEEEmembership{Senior Member,~IEEE}
\thanks{
A. Licciardi and L. Rondoni are with the Department of Mathematical Sciences ``G.L. Lagrange,'' Politecnico di Torino, Turin, Italy, and
Istituto Nazionale di Fisica Nucleare, Sezione di Torino, Turin, Italy (\texttt{\{alessandro.licciardi,lamberto.rondoni\}@ polito.it}).   R. Raineri, A. Proskurnikov, and L. Zino are with the Department of Electronics and Telecommunications,
Politecnico di Torino, Turin, Italy (\texttt{\{roberta.raineri,anton.proskurnikov,lorenzo.zino\}@ polito.it}).
R. Raineri is now with Modelway, Turin, Italy. A.Licciardi  and L.Rondoni worked under the auspices of Italian National Group of Mathematical Physics (GNFM) of INdAM.  A.Licciardi was supported by the Project Piano Nazionale di Ripresa e Resilienza --- Next Generation EU (PNRR-NGEU) from Italian Ministry of University and Research (MUR) under Grant DM 117/2023.
L. Rondoni  gratefully acknowledges support from the Italian Ministry of University and Research (MUR) through the grant PRIN2022-PNRR project (No. P2022Z7ZAJ) ``A Unitary Mathematical Framework for Modelling Muscular Dystrophies" (CUP: E53D2301807 0001). This manuscript reflects only the authors’ views and opinions, neither the European Union, the European Commission, nor the Ministry can be considered responsible for them.
}}
\maketitle
\thispagestyle{plain}
\pagestyle{plain}
\begin{abstract}
Server-based federated learning (FL) enables privacy-preserving collaborative model training, but its effectiveness is often constrained by heterogeneity in client data distributions.
In this paper, we address this limitation by introducing a client selection algorithm that i) dynamically forms non-overlapping coalitions of clients based on asymptotic agreement and ii) selects one representative from each coalition to minimize the variance of model updates.
Our approach is inspired by social network modeling, leveraging homophily-based proximity matrices for spectral clustering and informative node selection for estimating a group's aggregate opinion.
We provide theoretical insights on convergence for the algorithm under standard FL assumptions. \textcolor{black}{Empirical validation against a broad set of heterogeneity-aware baselines demonstrates that our method achieves superior or highly competitive accuracy under non-IID data distributions.}
\end{abstract}
\section{Introduction}
\IEEEPARstart{F}{ederated} learning (FL) has emerged as a powerful paradigm for enabling collaborative model training across multiple agents without directly sharing sensitive local data~\cite{mcmahan2017communication}. By decoupling model optimization from data aggregation, FL offers a principled framework for privacy-preserving and regulation-compliant machine learning, which is particularly appealing in data-sensitive domains such as healthcare~\cite{Yu2025_health, Yang2025_health}, finance~\cite{Fang2025_finance}, and IoT applications~\cite{Wong2025_IoT}. As a result, FL attracts significant and growing interest from the scientific community, spurring research on communication efficiency, robustness, personalization, and security in distributed learning systems. In the server-based FL setting considered in this work, a central aggregator coordinates training across distributed clients, enabling efficient model aggregation and scalable knowledge sharing while preserving the client data privacy.
\textcolor{black}{IoT ecosystems constitute a particularly compelling application domain for the framework developed in this work. Federations of smart devices and edge gateways operate under tight uplink-bandwidth and energy budgets, communicate over unreliable wireless links, and collect heterogeneous data, since each device senses a specific user, location, or physical process~\cite{Wong2025_IoT,kairouz2021advances}. In such systems, partial participation is a structural constraint rather than a design choice: only a limited number of devices can upload their model updates in each communication round~\cite{bonawitz2019towards}. Moreover, in view of the per-round uplink traffic and the energy consumption~\cite{li2019smartpc}, the effectiveness of an FL deployment ultimately hinges on how much information the server extracts from each transmitted update. Henceforth, by grouping statistically similar clients into coalitions and sampling from each coalition its most informative representative, the proposed method maximizes the statistical value of a fixed communication budget, thereby reducing the number of rounds required to attain a target accuracy and, with them, the aggregate bandwidth and energy consumption. 
}
\subsection{Literature review}
While FL reduces the need to centralize data, improving privacy and regulatory compliance, its effectiveness is often constrained by statistical heterogeneity among clients. In practice, client data are often non-IID, with each dataset following a distinct distribution~\cite{li2020federated,Lu2024}. This heterogeneity can significantly hinder the performance and convergence of the global model, leading to inconsistent local updates, slower convergence, and reduced model generalization, thus undermining the potential benefits of collaborative learning.
To address this challenge, several approaches have been proposed. Regularization methods, such as \texttt{FedProx}~\cite{li2020federated}, \texttt{FedMD}~\cite{li2019fedmd}, \texttt{Scaffold}~\cite{karimireddy2020scaffold}, and \texttt{Mime}~\cite{karimireddy2020mime}, constrain local client updates or modify the global aggregation process to improve robustness and convergence in heterogeneous settings.
Personalization techniques adapt the global model or learn client-specific models tailored to individual data distributions~\cite{smith2017federated, finn2017model}, with notable examples including \texttt{Per-FedAvg}~\cite{fallah2020personalized} and \texttt{pFedMe}~\cite{t2020personalized}. Collaboration and clustering strategies identify and leverage similarities among clients by grouping them into clusters, enabling more effective federated training within these clusters. Prominent approaches in this domain include \texttt{IFCA}~\cite{ghosh2020efficient}, \texttt{CFL}~\cite{sattler2020clustered}, \texttt{FeSEM}~\cite{long2023multi}. More recent studies on clustered FL have explored forming groups via consensus-based optimization~\cite{carrillo2024fedcbo} or by training pairwise discriminators to estimate client similarities~\cite{bao2023optimizing}.
Concurrently, client-sampling strategies have gained prominence. Practical issues such as intermittent client availability, energy constraints, and limited bandwidth necessitate selecting only a small subset of clients in each round~\cite{bonawitz2019towards}. While uniform  sampling serves as a common baseline, growing evidence suggests that a biased selection, taking data heterogeneity into account, can accelerate convergence and improve model quality~\cite{cho2022towards,goetz2019active}.  Various methods address data heterogeneity by prioritizing clients with higher local loss on the current global model~\cite{cho2022towards} or by promoting diverse client participation~\cite{balakrishnan2022diverse}. Other approaches focus on variability in client training dynamics~\cite{diao2020heterofl}, data and/or update quality~\cite{liao2024quality}, or energy efficiency~\cite{li2019smartpc}. Methods such as \texttt{FedCBS}~\cite{zhang2023fedcbs} prioritize clients to achieve more class-balanced sampling in each training round. More advanced policies, such as \texttt{Oort}~\cite{lai2021oort} and \texttt{Harmony}~\cite{tian2022harmony}, employ analytical rules to address heterogeneity in device capabilities and resources. A distinct research thread develops selection policies with reinforcement learning, typically framing client selection as a Markov decision process~\cite{powell2021reinforcement}. Notable examples include \texttt{FedRank}~\cite{tian2024fedrank}, \texttt{AutoFL}~\cite{kim2021autofl}, \texttt{Favor}~\cite{wang2020optimizing}, and \texttt{FedMarl}~\cite{zhang2022multi}.
In recent years, interest in FL has extended beyond the machine learning community to other disciplines. Game theory provides a rigorous framework for modeling strategic incentives and collaboration among federated learning participants. Early work~\cite{guazzone2013distributed} applied coalition formation games to energy-aware resource management in distributed systems, laying the groundwork for strategic collaboration, though not specifically addressing FL's statistical challenges. Donehue and Kleinberg used hedonic games to analyze coalition stability in federated linear regression, deriving MSE-optimal aggregation schemes, but limited their analysis to linear models~\cite{donahue2021model}. This was extended in~\cite{donahue2021optimality}, which framed \texttt{FedAvg} as a coalition game, introducing the price of anarchy to quantify inefficiencies in naive collaboration. The optimal coalition formation algorithm has limited scalability, though submodularity of the cost function suggests potential for efficient approximations. Blum et al.~\cite{blum2021one} studied equilibria in different FL collaboration structures (\textit{one-for-one} vs. \textit{all-for-all}), characterizing when each is optimal (under stylized utility models). To address dynamic interactions, Ota et al.~\cite{ota2022coalitional} proposed graphical coalitional games, forming coalitions via synergy measures such as cosine similarity or Improvement Classification Accuracy and analyzing robustness under adversarial settings. Nagalapatti et al.~\cite{nagalapatti2021game} proposed \texttt{S-FedAvg} for incentive-aligned client selection, using Shapley values to quantify client utility and exclude unhelpful participants.
It is worth noticing that the client selection problem is related to a broad class of \emph{subset selection} problems that have been extensively studied in various contexts and spanning a wide range of domains. Examples include feature selection in machine learning~\cite{Kempe,KempeApproxSubmod,5693974}, sensor placement for environmental monitoring~\cite{Krause2}, smart testing problem
to contain the spread of epidemics~\cite{smarttesting}. A greedy algorithm for selecting the most informative agents in an opinion dynamics framework was proposed in~\cite{raineri2025_ECC_ext, raineri2023_ifac}.
\subsection{Contribution}
In this paper, we introduce a novel client sampling algorithm termed \shortname\,(Federated Coalition Variance Reduction with Boltzmann Exploration) that addresses statistical heterogeneity across client distributions via a two-step procedure, as illustrated in Figure \ref{fig:method_overview}. The first step is an adaptive coalition detection phase that dynamically groups clients with similar model states. The second step is a within-coalition selection process that uses a Boltzmann-like probability measure to maximize the expected intra-coalition variance reduction~\cite{gallavotti2013statistical}, thereby directly mitigating the effect of heterogeneity.
\textcolor{black}{Our main focus is conceptual and methodological: we position client selection in FL through an analogy with homophily-based social selection mechanisms~\cite{Mas2010_opinion_exp}, and we integrate established tools, i.e., spectral clustering, similarity-based grouping, Boltzmann exploration, and variance-reduction sampling, into a unified selection strategy. We leverage results from estimation theory~\cite{kay1993} to justify the variance-reduction principle underlying our algorithm, building on related insights developed for opinion dynamics in~\cite{raineri2025_ECC_ext} and adapting them to the FL setting. Thus, the novelty of the framework does not lie in its individual methodological components, but in their principled integration through a social-network perspective, which provides a unified interpretation of client similarity, coalition formation and informative client selection.  In summary:
\begin{itemize}[leftmargin=*]
\item[i)] We propose \shortname, a client-sampling algorithm for heterogeneous FL that combines homophily-inspired client similarity, spectral clustering and variance-reduction sampling within a unified framework. Drawing on social-interaction models, we interpret groups of statistically similar clients as coalitions and select in each coalition the most informative client for the global model update.
\item[ii)] We provide theoretical insights into the convergence properties of the proposed algorithm under mild assumptions, and we characterize its computational complexity.
\item[iii)] We extensively evaluate \shortname\ across different heterogeneous settings and standard FL benchmarks~\cite{caldas2018leaf}, showing that the proposed strategy can outperform existing client-sampling baselines under non-IID data distributions.
\end{itemize}}
\textcolor{black}{It is worth clarifying the operating regime for which \shortname\ is designed. Following the standard taxonomy of FL~\cite{kairouz2021advances}, our method primarily targets the \emph{cross-silo} and moderate-scale IoT setting, in which the federation comprises from a few tens up to a few hundreds of clients, such as institutions, organizations, or edge gateways, rather than the extreme \emph{cross-device} regime involving up to millions of participants. In this regime, the cost of forming and clustering a client-similarity matrix (which is common practice in clustered FL~\cite{sattler2020clustered,vahidian2023efficient}) is negligible and, crucially, is incurred \emph{entirely on the server}, which is assumed to be computationally unconstrained: the clients only perform local training and transmit their updates, so that no additional computation, energy, or communication burden is placed on the resource-constrained edge devices. For larger federations, the quadratic cost can be further amortized by keeping the coalitions fixed over several communication rounds before re-clustering, consistently with the thermalized regime exploited in our analysis in Section~\ref{sec:theoretical_results}; the stochasticity of the Boltzmann sampling preserves within-coalition exploration and prevents the repeated selection of the same client. The same stochasticity makes the policy robust to intermittent device availability, since restricting the sampling measure to the currently reachable clients of each coalition preserves its structure; while approximate spectral methods such as the Nystr\"om approximation~\cite{fowlkes2004spectral} extend applicability to large-scale cross-device deployments.}
The rest of the paper is organized as follows. Section~\ref{sec:Framework} introduces the FL framework and the problem statement.
A detailed presentation of the algorithm and its pseudocode is given in Section~\ref{sec:algorithm}.
The core theoretical insights into our algorithm are presented in Section~\ref{sec:theoretical_results}.
In Section \ref{sec:experiments}, we present the numerical simulations. Section~\ref{sec:conclusion} concludes the paper and outlines future research directions.
\section{Federated Learning}\label{sec:Framework}
 The standard FL framework~\cite{mcmahan2017communication} involves $K$ clients (agents), each holding its own training data. Over $T$ communication rounds, the clients jointly estimate a global vector of $D$ parameters $\theta_{gl}\in\mathbb{R}^D$, often  called the global model. Ideally, the estimate should minimize the global loss:
\begin{equation}
  \theta_{gl}\in \arg\min\nolimits_{\theta \in \mathbb{R}^D} \mathcal{L}(\theta),
\end{equation}
where $\mathcal{L}(\theta) := \sum_{k=1}^{K} \frac{n_k}{n} \,\mathcal{L}_k(\theta)$, with $\mathcal{L}_k(\theta)$ denoting the client-level loss  computed on client $k$’s local dataset, where $n_k$ denotes the number of training samples held by the $k$-th client and $n = \sum_{k = 1}^K n_k$.
\vspace{1em}
\subsubsection*{Weighted-average global model} We adopt the standard Federated Averaging (FedAvg) framework~\cite{mcmahan2017communication}. Unlike ensemble methods that simply average final local solutions -- which would fail to minimize the global objective $\mathcal{L}(\theta)$ -- FedAvg minimizes the global loss iteratively.
Starting from a random initialization of the global model, at each round $t$, the server sends the current global model $\theta_{gl}(t)$ to the clients. Then, each client performs $S\geq 1$ local training iterations using a stochastic optimizer (e.g., stochastic gradient descent) to minimize their local loss
$\mathcal{L}_k$, thus resulting in a locally updated model $\theta_k(t+1)$. Each client communicates the updated model back to the server, which aggregates the updates as a weighted average, namely
\begin{equation}\label{eq:global_model}
    \theta_{gl}(t+1)=\sum\nolimits_{k=1}^K \alpha_k(t) \theta_k(t+1).
\end{equation}
where the weights $\alpha_k(t)$ are such that $\sum_{k =1}^K \alpha_k(t) = 1$. In general, as in \cite{mcmahan2017communication}, they are set $\alpha_k(t)  = \frac{n_k}{n}$ for every $t=1,\dots,T$. However, other algorithms propose different weighting schemes, e.g., \cite{wang2020tackling}.
By periodically resetting clients to the shared global state,~\eqref{eq:global_model} effectively averages the accumulated gradients across clients. Theoretical analysis confirms that this iterative process converges to a stationary point of the global loss function $\mathcal{L}(\theta)$, rather than a suboptimal average of local minima \cite{wang2020tackling}.
This setting is natural when all clients aim to estimate the same $D$-dimensional parameter vector using only their local datasets, under the assumption that all clients share the same model architecture. 
\subsubsection*{Partial participation} Since the number of clients $K$ can be very large, the server typically interacts with only a fraction of them in each round. This partial-participation strategy keeps both the throughput and energy consumption manageable, mitigates stragglers, and thereby shortens each communication round without compromising convergence~\cite{mcmahan2017communication,Luo2024}. Specifically, the global model is communicated to the randomly selected subset of clients $\mathcal{P}_t\subseteq\{1,\dots,K\}$ whose size $P=|\mathcal{P}_t|$ is kept constant across rounds \textcolor{black}{and coincides with the maximum number of clients that can participate in each communication round, thus reflecting the system’s available communication and computational resources.} \textcolor{black}{We remark that $P$ also determines the number of coalitions into which clients are grouped (see Section~\ref{sec:algorithm}), so that $P/K$ coincides with the participation rate and $K/P$ with the average coalition size. The method is therefore naturally suited to the limited-participation regime typical of realistic FL deployments: excessively large participation rates would induce over-clustering, with coalitions shrinking to one or two clients, which both undermines the variance-reduction rationale and exposes individual clients; a sensitivity analysis on $P$ is reported in Section~\ref{sec:ablation}.}
Only these selected clients compute their local updates $\theta_k(t+1)$ and upload them to the server for aggregation.
The server updates the global model by aggregating the received estimates:
\begin{equation}
  \theta_{gl}(t+1)  =
  \sum\nolimits_{k \in \mathcal{P}_t} \tilde{\alpha}_k(t)\,\theta_k(t+1).
\end{equation}
where the weights $\tilde{\alpha}_k(t)$ satisfy $\tilde{\alpha}_k(t)\ge 0$ and $\sum_{k\in\mathcal{P}_t}\tilde{\alpha}_k(t)=1$. These weights differ from the original $\alpha_k$ as they are re-normalized to the active subset $\mathcal{P}_t$.
\subsubsection*{Statistical learning setup} We now assume that each client’s training set is a sample drawn from an underlying data-generating distribution. The parameter vector $\theta_k\in\mathbb{R}^D$ learned by client $k$ (typically, obtained by minimizing its local empirical risk) thus becomes a random variable. Consequently, the vector $\theta_k(t)$ produced by client $k$ in communication round $t$ is a realization of that random variable, which the server then incorporates into the global aggregation. We define the random vector $\theta^d = (\theta_1^d,\dots,\theta_K^d)^\top \in \mathbb{R}^K$,
whose entries are the $d$-th components of the clients’ parameter vectors, with the mean
$\bar{\theta}^d = \mathbb{E}\bigl(\theta^d\bigr)$
and the symmetric covariance matrix
$C^d = \cov \bigl(\theta^d\bigr) \in \mathbb{R}^{K\times K}.$
The entries of $C^d$ quantify client heterogeneity for the $d$-th model parameter: its diagonal elements capture the variance across clients, while the off-diagonal entries measure pairwise similarity between clients’ values. 
Consistent with the aggregation logic of the estimators at communication round $t$ in \eqref{eq:global_model}, it is natural to define each global model components as a scalar random variable corresponding convex combination of the local model random variables $d$-components, i.e., $\theta_{gl}^d = \alpha^{\top}\theta^d$.
We make a standard assumption on the covariance matrix that will be useful in the subsequent analysis and is not restrictive since the diagonal elements refers to clients' variance.
\begin{assumption}
    The matrix $C^d$ has positive diagonal elements, i.e., $C^d_{kk}>0$ for every $k$.
\end{assumption}
\subsubsection*{Heterogeneity in FL}
Recent studies show that, under data heterogeneity, the choice of the participating-client set $\mathcal{P}_t$ is  critical  for convergence as biased sampling policies can significantly accelerate and stabilize training~\cite{ghosh2020efficient,cho2022towards}. In the rest of this paper, we provide a theoretically-principled approach to address heterogeneity. We introduce a novel FL algorithm with a client selection mechanism inspired by social dynamics models, which performs biased sampling of clients. We demonstrate the algorithm's performance both theoretically and empirically through numerical studies against established benchmarks.
We summarize the notation in Table~\ref{tab:placeholder}.
\begin{table}[t]
\centering
\footnotesize
\setlength{\tabcolsep}{1pt}
\caption{Notation summary. }\begin{tabular}{>{\raggedright\arraybackslash}p{0.2\linewidth} >{\raggedright\arraybackslash}p{0.79\linewidth}}
\rowcolor{gray!10}\textbf{Symbol} & \textbf{Description} \\
\hline
\(K\) & number of clients \\
\(\mathcal{K}\) & set of clients \\
\(W(t)\) & influence matrix among clients at round \(t\) \\
\(T\) & number of communication rounds \\
\(D\) & number of model parameters \\
\(S\) & number of training iterations \\
\(P\) & number of participating clients \\
\(\eta\) & learning rate \\
\(\theta_{{gl}} \in \mathbb{R}^D\) & global model \\
\(\theta_{{gl}}(t) \in \mathbb{R}^D\) & global model at round \(t\) \\
\(\theta_k(t) \in \mathbb{R}^D\) & local model of client \(k\) at round \(t\) \\
\(\tilde \theta_k(t) \in \mathbb{R}^D\) & normalized local model of client \(k\) at round \(t\) \\
\(\mathcal{L}(\theta)\) & global loss \\
\(\mathcal{L}_k(\theta)\) & client loss \\
\(\mathcal{P}_t\) & subset of participating clients at round \(t\) \\
\(\alpha_k\) & aggregation weight of client $k$ in the full federation\\
\(\tilde{\alpha_k}(t)\) & aggregation weight of $k$ at round $t$, re-normalized over \(\mathcal{P}_t\)\\
\(\theta_{{gl}}^{d} \in \mathbb{R}^{K}\) & random vector of \(d\)-th components of the global model \\
\(\theta^{d} \in \mathbb{R}^{K}\) & random vector of \(d\)-th components of the model \\
\(\bar{\theta}^{d}\) & mean of random vector \(\theta^{d}\) \\
\(\bar{\theta}_{k}(t+1)\) & mean of unobserved clients $k \in \{1,\dots,K\}\backslash \mathcal{C}_i(t)$ \\
\(C^{d}\) & covariance of random vector \(\theta^{d}\) \\
\(\rho_{jk}\) & Pearson correlation between random variables \(j\) and \(k\) \\
\(v^{d} \in \mathbb{R}^{K}\) & value vector for the \(d\)-th component; \(v_k^{d}\) is the variance reduction for estimating \(\theta_{\mathrm{gl}}^{d}\) when sampling \(\theta_k^{d}\) \\
\(v_k\) & total variance reduction associated with client \(k\) \\
\(\Pi(K)\) & set of all partitions of \(\{1,\dots,K\}\) \\
\(\mathcal{C}_i(t) \in \Pi(K)\) & \(i\)-th coalition at round \(t\) \\
\(\pi_p(k;t)\) & Boltzmann-like probability that client \(k\) is selected in \(\mathcal{C}_i(t)\) \\
\(n_k\) & number of training samples of $k$-th client\\\hline
\end{tabular}
\label{tab:placeholder}
\end{table}

\section{Algorithm Design} \label{sec:algorithm}

\begin{figure}[t]
    \centering
    \includegraphics[width = \linewidth]{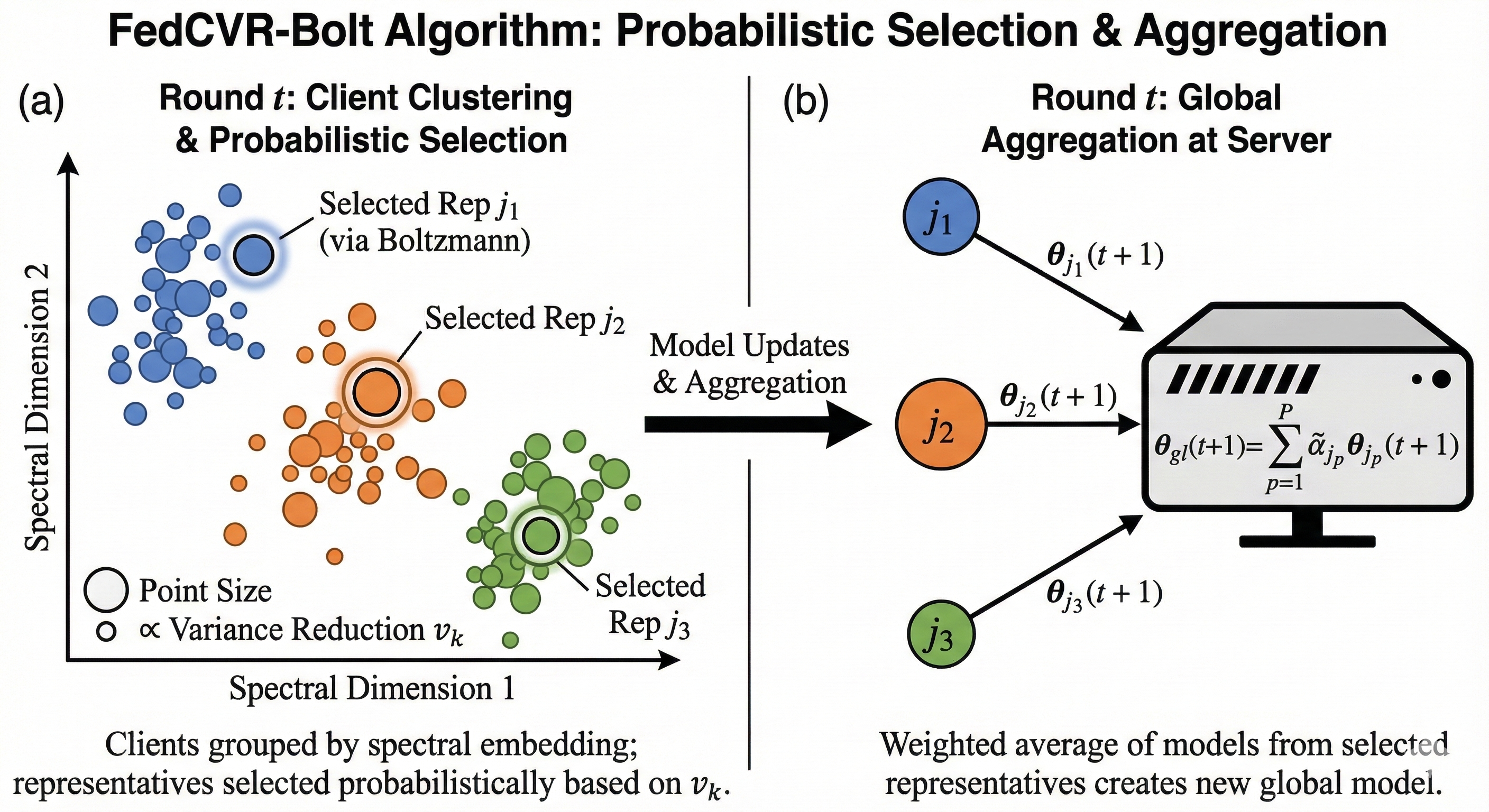}
    \caption{ \small representation of \shortname{} sampling and aggregation strategy at round $t$. \textit{Left.} Clients are grouped according to the affinity of the models distribution into $P$ clusters. Colors denote the clusters ($\mathcal{C}_1, \dots, \mathcal{C}_P$) identified via spectral clustering. The size of each node is proportional to its variance reduction $v_k$, representing the informative value of the client's model. One representative is sampled from each cluster using the Boltzmann-like probability $\pi_p(k;t) \propto e^{v_k}$. \textit{Right.} The selected updates $\theta_{j_p}(t+1)$ are aggregated by the server to compute the next global model $\theta_{gl}(t+1)$.}
    \label{fig:method_overview}
\end{figure}

As discussed in the previous sections, a key challenge in FL is to design an optimal partial participation taking into consideration data heterogeneity. Based on the state-of-the-art literature and building on the statistical learning setup previously introduced, we introduce a novel sampling strategy based on variance reduction techniques. Precisely, we will partition the federation into $P$ groups of cooperating clients whose data are drawn from the same (or closely related) distribution, and then, at each round, we will select the best representative from each group using the strategy detailed below.
In the following, we present the steps performed at every communication round $t \in \{1,\dots,T\}$, illustrated in Fig.~\ref{fig:method_overview}. Then, we summarize the procedure in Algorithm \ref{alg:fed_cluster}.

\subsubsection*{Analogy with Social Networks} We formalize the underlying network structure aimed at studying the problem through the lens of social agents. Here, the set of clients of the federations can be seen as the set of agents $\mathcal{K} = \{1, \dots, K\}$ of an arbitrary network and their individual client model $\theta_k$ as the agents' social opinions~\cite{proskurnikov2017tutorial}. Let $\theta(t)$ indicate the vector of current models at round $t$, such that $\theta_k(t)$ is the model associated to client $k$ at communication round $t$. Let $W(t)$ be the influence matrix that encodes the pairwise influence between clients at time $t$. In the context of opinion dynamics, this matrix captures phenomena such as homophily~\cite{Mas2010_opinion_exp}, i.e., agents' tendency to be mainly influenced by individuals who hold  similar opinions. 
Following~\cite{Mas2010_opinion_exp}, a natural way to implement homophily is to define social influence of agent $j$ on agent $k$ as 
\begin{equation}
    W_{kj}(t) = \frac{e^{- \gamma \|\theta_k(t)-\theta_j(t)\|_2^2}}{\sum_{j \in \mathcal{K}} e^{- \gamma \|\theta_k(t)-\theta_j(t)\|_2^2}},
\end{equation}
where $\gamma>0$ is a parameter that amplifies the role of models similarities, i.e., the larger $\gamma$, the more pronounced is the homophily effect.
The resulting matrix admits a clear social-systems interpretation: agents (clients) with similar opinions (model parameters) exert stronger mutual influence and naturally form clusters. Importantly, computing this matrix, which is standard practice in clustered and personalized FL~\cite{sattler2020clustered, bao2023optimizing}, here reduces to evaluating  
Euclidean distances between server-available parameter vectors.
It is worth noting that we expect clients belonging to the same cluster to converge toward similar local models. Consequently, the influence matrix $W$ will asymptotically exhibit a block-like structure, with each block corresponding to a distinct cluster. This observation supports the specific formulation of $W$, as it aligns with our objective of accurately identifying and distinguishing network clusters.

\subsubsection*{Coalition Formation}
We move the focus on the coalitions formed which may vary with respect to the previous round. Formally, the goal of this step is to identify the $P$ coalitions formed at round $t$ that will be indicated by $\mathcal{C}_1, \dots, \mathcal{C}_P \subset \Pi(K)$, where $\Pi(K)$ denotes the set of all possible partitions of the set $\{1,\dots,K\}$. Notice that the number of communities $P$ is neither  assumed to be the true number of clusters in the data, nor is an output of the clustering step, but it is fixed a priori and it coincides with the model participation rate. This assumption is not restrictive since it is common practice in clustered framework (e.g., refer to IFCA~\cite{ghosh2020efficient} and FeSEM~\cite{long2023multi}). 

As proposed in~\cite{Narantsatsralt2017_spectralClust_OD}, a natural way to detect communities in social networks is via spectral clustering, as illustrated in the left panel of Fig.~\ref{fig:method_overview}. 
Following this approach, we project the data, collected in matrix $W$, onto the corresponding eigenvector space and then we apply a k-means on this space. The projection onto the eigenvectors space is key since it allows to significantly distinguish similar nodes into more distanced positions in feature space, assigning similar values to members of the same community and capturing  soft boundaries between communities, both convex and non-convex.  

\subsubsection*{Sampling Strategy}
Once the $P$ coalitions of clients have been identified, the key step of our algorithm consists in determining the optimal client to sample for each coalition. To do this, we adopt as target measure the variance reduction~\cite{raineri2025_ECC_ext}, which is known to evaluate the most informative nodes over a network, i.e., the best nodes to observe to minimize the variance on the complete knowledge about network status $\theta^d_{gl} $. The formal definition of the variance reduction vector is presented in the following proposition, which builds on~\cite[Proposition~2]{raineri2025_ECC_ext}, properly adapted to our specific framework.
Preliminary, notice that, given $\theta_{gl}^d$ the $d$-th component of the global model to estimate given $\theta_k^d$ the sampled model, the associated variance reduction is computed as
\begin{equation}\label{eqn:var_red}
\var(\theta_{gl}^d)-\mathbb{E}[\var(\theta_{gl}^d|\theta_{k}^d)] = \var(\mathbb{E}[\theta_{gl}^d|\theta^d_{k}]),
\end{equation}
where the second equality comes from law of total variance. 
{\color{black}
\begin{remark}
   For the sake of clarity in our exposition, we assume a zero-mean distribution of $\theta^d$. This does not limit the generality of our findings, since the measure of interest is translation invariant,  i.e., $\var(\theta_{gl}^d)-\mathbb E[\var(\theta_{gl}^d|\theta_{k}^d)] = 
    \var( \theta_{gl}^d- \bar \theta_{gl}^d)-\mathbb E[\var(\theta_{gl}^d - \bar \theta_{gl}^d|\theta_{k}^d - \bar \theta_{k}^d)] \,.$
\end{remark}
}
{While computing the conditional expectation can be difficult for non-Gaussian random variables, we can instead approximate it by the $L^2$-optimal linear projection of $\theta_{gl}^d$ onto $\theta^d_k$, that is,
\begin{equation}\label{eq:proj1}
\operatorname{Proj}(\theta_{gl}^d|\theta_{k}^d)=\frac{\mathbb{E}[\theta_{gl}^d\theta_{k}^d]}{\mathbb{E}[(\theta_{k}^d)^2]}\theta_{k}^d
\end{equation}
This projection is equivalent to the standard least-squares linear regression of $\theta_{gl}^d$ on $\theta_k^d$ (with zero intercept under the assumption of centered variables). The approximation\footnote{This approximation is exact when $\theta^d$ is Gaussian, since for jointly Gaussian random variables the conditional expectation $\mathbb{E}[\theta_{gl}^d \mid \theta_k^d]$ is affine (and reduces to a linear function when the variables are centered).} of the variance reduction~\eqref{eqn:var_red} can then be defined as
\begin{equation}
v_k^d:=\var(\operatorname{Proj}(\theta_{gl}^d|\theta_{k}^d))=\frac{\big(\mathbb{E}[\theta_{gl}^d\theta_{k}^d]\big)^2}{\mathbb{E}[(\theta_{k}^d)^2]}.
\end{equation}
Applying this to the global model $\theta^d_{gl} = \alpha^{\top}\theta^d$ and by noticing that $\theta_k^d=\mathbf{e}_k^{\top}\theta^d$, where $\mathbf{e}_k$ is the $k$-th canonical basis vector, the (approximate) variance reduction $v_k^d$ is thus found as follows:
\begin{equation}\label{eqn:var_red+}
        v_k^d = \frac{(\mathbf{e}_k^{\top}C^d\alpha)^2}{\mathbf{e}_k^{\top}C^d\mathbf{e}_k}=\frac{(C^d\alpha)_k^2}{C^d_{kk}}.
 \end{equation}
 }



So far, we have proposed an explicit formulation for the variance reduction function of the global model $\theta_{gl}^d$ for each $d$-th component, for $d = 1,\dots, D$. Now, we define a collective measure of the overall variance reduction of $\theta_{gl}$. Based on~\cite{johnson2002_ult_statistical_anal}, we define the {total variance reduction} as the sum of variance reductions associated to the different model components, i.e.,
\begin{equation}
    v_k = \sum\nolimits_{d=1}^D v_k^d,
\end{equation}
which approximates the trace of $\var(\mathbb{E}(\theta_{gl}|\theta_{k}))$, coinciding with the sum of the $D$ squared deviation vectors.

  The introduced centrality measure allows us to define an optimal sampling strategy. Precisely, for each coalition we select the client maximizing vector $v$, i.e., for each coalition $p = 1,\dots,P$ we select the client $j_p=j_p(t)$ such that
\begin{equation}\label{eq:selection}
j_p \in \arg\max\nolimits_{k \in \mathcal{C}_p(t)} v_k  \,.  
\end{equation}
{Selecting one representative per cluster mitigates overfitting to any single sub-distribution while preserving global diversity. Optimizing for variance reduction captures cross-client coupling, yielding more stable updates.}

\subsubsection*{Boltzmann Exploration}
It is worth noticing that, in real-world scenarios, the optimization in \eqref{eq:selection} could become  restrictive due to the unavailability of the true data distribution, which introduces an inherent estimation error in the covariance matrix.  To mitigate the impact of this uncertainty, while exploiting the informative content of the estimated variance, we incorporate an additional exploration term into the objective.

In particular, instead of a simple greedy, deterministic selection based on the estimated variance, we employ a Boltzmann exploration policy~\cite{gallavotti2013statistical}, as illustrated in the left panel of Fig.~\ref{fig:method_overview}. This introduces a controlled stochasticity that helps mitigate the risk of becoming trapped by early biased or noisy estimates of the covariance matrix~\cite{Powell2022ReinforcementLA}.  Drawing an analogy with statistical mechanics~\cite{gallavotti2013statistical}, where the probability of a system occupying a specific energy state is proportional to the exponential function of the negative energy divided by temperature, we let the probability of selecting a client $k$ within coalition $\mathcal{C}_p(t)$ during round $t$ based on its associated value $v_k$ be an indicator of the desirability of sampling client $k$. Higher values indicate greater potential to reduce the variance of the global model estimate.
This Boltzmann-like probability measure is defined as:
\begin{equation}\label{eq:boltzmann_probs}
\pi_p(k;t) = \dfrac{e^{ v_k}}{\sum_{j \in \mathcal{C}_p(t)}e^{ v_j}},
\end{equation}
where $\pi_p(k;t)$ is the probability that client $k$ in coalition $\mathcal{C}_p(t)$ is selected at round $t$. This design assigns clients with larger estimated variance-reduction $v_k$ proportionally higher selection probabilities, aligning sampling with their expected contribution within each coalition. Moreover, this framework keeps nonzero probability for lower-scoring clients, preventing myopic exploitation and enabling gains under uncertainty.
\color{black}{In particular, $\pi_p(k;t)$ is strictly positive, so that no client can be permanently excluded from selection; together with the uniform-sampling warm-up phase (Appendix~\ref{app:exp}), this guarantees a balanced coverage of the client population.}
\subsubsection*{Update Policy}
Once the participating clients $\mathcal{P}_t = \{j_1, \dots, j_P\}$ have been selected, the server sends them the current model. Then, they locally update the central model according to their local dataset, obtaining the individual update $\theta_{j_p}(t+1)$ which is sent to the server. Finally, it coherently updates the global model as \begin{equation}\theta_{gl}(t+1) = \sum\nolimits_{p = 1}^P \tilde{\alpha}_{j_p}(t) \theta_{j_p}(t+1),
\end{equation}
as illustrated in the right panel of Fig.~\ref{fig:method_overview}.

For the unobserved clients $k \in [K]\backslash \mathcal{C}_p(t)$, the value at round $t+1$ is considered equal to the previous one at $t$, i.e., $\theta_k(t+1) = \theta_{j_p}(t+1)$ if $k=j_p$, and $\theta_k(t+1) = \theta_k(t)$ if $ k\neq j_p$. 
To properly estimate the updated conditioned covariance matrix that will be used in the new round, an update policy for the expected value of $\theta_k$ must be formalized. 
\begin{proposition} \label{prop:update}
   Let $\theta_{j_p}(t+1)$ be the sampled model at round $t+1$. The best linear predictor (in the mean squared error sense) of the $d$-th component
   $ \theta^d_k$ for client $k \neq j_p$, given the observation $\theta_{j_p}(t+1)$, is
   \begin{equation}\label{eq:observation}   
   \bar \theta_k^d(t+1) =\tilde \rho_{kj_p}^d(t)\theta^d_{j_p}(t+1),
   \end{equation}
for all components $d$, where the linear regression coefficient $\tilde \rho_{kj_p}(t)$ is obtained via normalization of the Pearson correlation coefficient between the models of clients $k$ and $j_p$, that is,
\[
\tilde \rho^d_{kj_p}(t) = \rho^d_{kj_p}(t)\tfrac{\sqrt{C^d_{kk}(t)}}{\sqrt{C^d_{j_pj_p}(t)}},\;\;\rho_{k j_p}^d(t)=\tfrac{C^d_{kj_p}(t)}{\sqrt{C^d_{kk}(t)}\sqrt{C^d_{j_pj_p}(t)}}.
\]
\end{proposition}
\begin{proof}
Omitting the time dependence for brevity, the linear projection of $\theta^d_k$ onto $\theta^d_{j_p}$ in the mean-square ($L_2$) sense is
    \begin{equation}\label{eq:proj2}
\bar \theta_k^d=\operatorname{Proj}(\theta_{k}^d|\theta_{j_p}^d)=\tfrac{\mathbb{E}[\theta_{j_p}^d\theta_{k}^d]}{\mathbb{E}[(\theta_{j_p}^d)^2]}\theta_{j_p}^d=\tfrac{C^d_{kj_p}}{C^d_{j_pj_p}}\theta_{j_p}^d.
\end{equation}
It remains to notice that from Pearson correlation coefficient definition it holds $\rho_{kj_p}=\frac{C^d_{kj_p}}{\sqrt{C^d_{kk}}\sqrt{C^d_{j_pj_p}}}$.
\end{proof}

\subsubsection*{Online estimation of covariance}
Final step is now the online estimation of the covariance matrix $C^d$ of the individual model's components which represent a central core of our method. Let us denote by $C^d(t)$ the current estimate at round $t$. Then, the following formulation holds
\begin{equation} \label{eqn:cov_def}
    C^d_{kj} = \cov(\theta_k^d, \theta_j^d) = \mathbb{E}[(\theta^d_k- \bar{\theta^d_k})(\theta^d_j- \bar{\theta^d_j})]\,.
\end{equation}


 Using the $k,j$-th entry of the covariance matrix in \eqref{eqn:cov_def}, it is possible to obtain an online estimate of the covariance matrix, using  the Robbins--Monro estimation~\cite{robbins1951stochastic, licciardiinteraction},   as 
    $C^d(t+1) = (1-\gamma_t) C^d(t)
   +\gamma_t[\theta^d(t+1)- \bar{\theta^d}(t+1)] [\theta^d(t+1)- \bar{\theta^d}(t+1)]^\top$,
with $\bar \theta^d(t+1)$ the mean vector updated according to \eqref{eq:observation}. 
\begin{assumption}
The sequence of update coefficients $\{\gamma_t\}_{t=1}^\infty$ satisfies $\gamma_t \in [0,1]$ for all $t\ge1$, and 
$\{\gamma_t\}_{t=1}^\infty \in \ell^2(\mathbb{N}) \setminus \ell^1(\mathbb{N})$,
i.e., $\sum\nolimits_{t=1}^\infty \gamma_t = \infty$ and 
$\sum\nolimits_{t=1}^\infty \gamma_t^2 < \infty$.
\end{assumption}

Under this assumption, the estimator $C^d(t)$ converges almost surely and in $L^2$ to the covariance matrix $\mathbb{E}\big[(\theta^d-\bar{\theta}^d)(\theta^d-\bar{\theta}^d)^\top\big]$,
see,~\cite{kushner2003stochastic, licciardiinteraction}. 
In our experimental evaluation we choose $\gamma_t = 1/t$, which satisfies the above conditions.
All the steps of the algorithm, extensively discussed in the above, are summarized in Algorithm~\ref{alg:fed_cluster}. 
\begin{algorithm}
\caption{\shortname~Algorithm}
\label{alg:fed_cluster}
\begin{algorithmic}[1]
\State \textbf{Inputs:} $K$, $D$, $T$, $\{\theta_k(1)\}_{k=1}^K$, $\{C^d(1)\}_{d=1}^D$, $\theta_{gl}(1)$, $\gamma_t$, $P$
\For{$t = 1,\dots,T-1$}
    \State Normalize: $\tilde{\theta}_k(t) = \theta_k(t)/\|\theta_k(t)\|$
    \State Compute influence matrix: $W_{kj}(t) = \dfrac{e^{-\gamma\|\tilde\theta_k(t)-\tilde\theta_j(t)\|_2^2}}{\sum_{m}e^{-\gamma\|\tilde\theta_k(t)-\tilde\theta_m(t)\|_2^2}}$
\State Set: $\{\mathcal{C}_i(t)\}_{i=1}^P \leftarrow \text{SpClustering}(W(t), P)$
    \State Compute $v_k^d$ with $C^d(t)$; set $v_k = \sum_{d=1}^D v_k^d$
    \State Compute $\pi_p(k;t) = {e^{v_k}}/{\sum_{j \in \mathcal{C}_p(t)} e^{v_j}}$
    \State Sample $j_p \sim \pi_p(\cdot;t)$ in each cluster $\mathcal{C}_p(t)$
    \State Set $\mathcal{P}_t = \{j_1,\dots,j_P\}$
    \ForAll{$j_p \in \mathcal{P}_t$}
        \State Receive $\theta_{gl}(t)$, perform local update $\rightarrow \theta_{j_p}(t+1)$
    \EndFor
    \ForAll{$p = 1,\dots,P$,  $k \in \mathcal{C}_p(t)$ and $d=1, \dots, D$}
        \State $\bar{\theta}^d_k(t+1) = \tilde \rho^d_{k j_p}(t) \theta_{j_p}^d(t+1)$
    \EndFor
    \For{$d = 1,\dots,D$}
        \State $C^d(t+1) = (1 - \gamma_t) C^d(t) + \gamma_t\left(\theta^d(t+1) - \bar{\theta}^d(t+1)\right) \left(\theta^d(t+1) - \bar{\theta}^d(t+1)\right)^\top$
    \EndFor
    \State $\theta_{gl}(t+1) =  \sum_{p} \tilde{\alpha}_{j_p}(t)\theta_{j_p}(t+1),$ $\tilde{\alpha}_{j_p}(t) = \frac{n_{j_p}(t)}{\sum_{p}n_{j_p}(t)}$
\EndFor

\end{algorithmic}
\end{algorithm}

\begin{remark}[Privacy Guarantees]
   Let us notice that the privacy of client data in \shortname \; algorithm is preserved by adhering to the fundamental principles of FL: raw data remains on client devices and is never transmitted to the server. Clients selected from the set $\mathcal{P}_t$ perform local updates using the global model $\theta_{gl}(t)$ to produce their updated local models $\theta_{j_p}(t+1)$. Moreover, clients subsequently communicate to the server only these model parameters, not the underlying private data. To further protect such parameters during aggregation steps (e.g., for the computation of $\theta_{gl}(t+1)$), it is possible to employ secure aggregation protocols~\cite{bonawitz2016practical}. While the distinctive server-side operations of \shortname, such as clustering based on individual models $\{\theta_k(t)\}$ and deriving $\bar{\theta}_k(t+1)$ from specific client model updates $\theta_{j_p}(t+1)$, require the server to access these individual parameters for its core functionality, this architectural choice is consistent with federated architectures where the central server orchestrates the learning process using the model parameters received from the clients~\cite{kairouz2021advances} for advanced tasks like personalization or clustering~\cite{smith2017federated}. 
\end{remark}

\section{Theoretical Analysis}\label{sec:theoretical_results}

\subsection{Convergence}

We derive insights into the convergence properties of \shortname. Our analysis builds on assumptions that are standard in the theoretical analysis of stochastic optimization and Federated Learning \cite{bottou2018optimization, liconvergence2020, wang2020tackling}: smoothness of the global loss and bounded variance of the stochastic federated gradient. We extend these with a mild alignment condition on the client selection policy.  
The latter ensures that, in expectation, the update direction preserves a positive correlation with the true gradient, thereby maintaining descent. Under these assumptions, the sequence of global iterates produced by our algorithm tends to approach a neighborhood of a stationary point of the global loss. In fact, as typically requested in the stochastic optimization sense, we are able to demonstrate that, once averaged over both the randomness of client sampling and the evolution in time, the gradient norm becomes arbitrarily small.  
Notice that this differs from  stronger notions of convergence of the iterates $\theta_{gl}$  to an exact stationary point, which is precluded by the intrinsic variance of stochastic updates but replaced by concentration around regions of vanishing gradient, but still provides insights into the convergent nature of our algorithm. {\color{black}In the following, we assume that the system  reaches a thermalized regime, where the covariance estimates and the induced sampling distribution approaches a stationary distribution, which is a  standard assumption in FL~\cite{kim24clustered}. In this regime, the coalitions and the sampling distribution are stationary; if the update noise has bounded second moment (see the assumption below), the covariance update also satisfies the standard stochastic-approximation conditions and the scheme is well posed. }


\begin{assumption}\label{a:1}
   The global loss function $\mathcal{L}(\theta)$ is $L$-smooth, i.e., it is differentiable with $L$-Lipschitz gradient and its federated gradient $g(t) := \sum_{k \in P_t}\tilde{\alpha}_k(t) \nabla \mathcal{L}_k(\theta_{gl}(t))$ {\color{black}has uniformly bounded second moment, i.e., $\mathbb{E}_{\mathcal{P}_t\sim \pi(t)}[\|g(t)\|^2] \leq M$, for some constant $M>0$.} {\color{black}Moreover, $\mathcal{L}$ is uniformly bounded from below, i.e., $\mathcal{L}^* := \inf_{\theta\in\mathbb{R}^D}\mathcal{L}(\theta) > -\infty$.}
\end{assumption}

\begin{assumption}\label{a:2}
 Let $G(\theta):= \mathbb{E}_{\mathcal{P}_t\sim \pi(t)}[g(t)]$. Assume there exists $c>0$ s.t. $ G(\theta)^\top \nabla \mathcal{L}(\theta) \geq c \|\nabla\mathcal{L}(\theta)\|^2$, i.e., the bias of the policy is informed and points in a descent direction.
\end{assumption}

\begin{theorem}\label{th:conv}
    Let $\{\theta_{gl}(t)\}$ be the sequence of global model update produced by FedCVR-Bolt algorithm. Under Assumptions~\ref{a:1}--\ref{a:2}, 
 there exists a finite constant $M>0$ such that, for a small learning rate $\eta>0$,
    \begin{equation}\limsup\nolimits_{T \to \infty} \dfrac{1}{T} \sum\nolimits_{t = 0}^{T-1} \mathbb{E}[\|\nabla\mathcal{L}(\theta_{gl}(t))\|^2]\leq {\color{black}\frac{LM\eta}{2c}}.\end{equation}
\end{theorem}
\begin{proof}
    The proof is reported in Appendix~\ref{app:th-conv}.
\end{proof}

{\color{black}Note that the two assumptions are key for the proof of the theorem, but are quite standard and realistic. In particular, Assumption~\ref{a:2} is standard in the analysis of stochastic-gradient methods with biased updates~\cite{bottou2018optimization, ajalloeian2020convergence}, and Theorem~\ref{th:conv} is conditional on it. It is worth noticing that Theorem~\ref{th:conv} is a general convergence result for biased stochastic gradients and does not provide an explicit characterization of the impact of the parameters of \shortname on the convergence rate, as they are indirectly in the parameters $M$, $L$, and $c$ from Assumptions~\ref{a:1}--\ref{a:2}. }

\subsection{Computational Cost}
The main overhead lies in performing spectral clustering on the server at each communication round, whose computational complexity can be computed as follows. 
\begin{proposition}
    The computational cost is $\mathcal{O}(K^2 D + K^2 P)$.
\end{proposition}
\begin{proof}
    The computational cost at each round can be broken down as follows~\cite{luxburg2007}. 
    First, for the similarity matrix construction, pairwise similarities between the $K$ clients are computed using a RBF kernel. This requires evaluating squared Euclidean distances between $K$ vectors in $\mathbb{R}^D$, resulting in a complexity of $\mathcal{O}(K^2 D)$. 
    Second, constructing the normalized Laplacian from the similarity matrix incurs a cost of $\mathcal{O}(K^2)$. 
    Third, extracting the top $P$ eigenvectors of the $K \times K$ Laplacian requires $\mathcal{O}(K^2 P)$ operations in standard implementations.
    Fourth, the final step clusters the  $K$ clients in the $P$-dimensional eigenspace, with cost $\mathcal{O}(KPI)$, where $I$ is the number of K-means iterations, yielding the claim. 
\end{proof}
 In practice, since $P \ll D$ (e.g., in our numerical setting, $P = 10$, $D = 300$), the complexity is effectively driven by the similarity matrix computation, yielding a leading cost of $\mathcal{O}(K^2 D)$ per round. This overhead is modest and entirely server-side, since server resources are not a bottleneck.

\section{Numerical Simulations}\label{sec:experiments}
\begin{table*}
\centering
\caption{\small Comparison with FL baselines on heterogeneous classification benchmarks. }
\label{tab:class}
\begin{adjustbox}{width = \textwidth}
\begin{tabular}{l|cccccccc>{\columncolor{gray!10}}c}
\toprule
\textbf{Dataset} & \texttt{FedAvg} & \texttt{Power-of-Choice} & \texttt{Active FL} & \texttt{FedProx} & \texttt{IFCA} & \texttt{FeSEM} & \textcolor{black}{\texttt{FedCBS}} & \textcolor{black}{\texttt{Oort}} & \shortname \\
\midrule
MNIST     & 86.30 \scriptsize{$\pm$ 1.12} & 81.00 \scriptsize{$\pm$ 2.10} & 88.89 \scriptsize{$\pm$ 0.98} & 89.69 \scriptsize{$\pm$ 0.91} & 89.46 \scriptsize{$\pm$ 1.04} & 86.77 \scriptsize{$\pm$ 1.85} & \textcolor{black}{90.05  \scriptsize{$\pm$ 3.07}} & \textcolor{black}{89.43 \scriptsize{$\pm$ 1.49}} & \textbf{90.23 \scriptsize{$\pm$ 0.82}} \\
CIFAR-10  & 52.67 \scriptsize{$\pm$ 1.25} & 49.85 \scriptsize{$\pm$ 1.71} & 55.19 \scriptsize{$\pm$ 1.34} & 46.44 \scriptsize{$\pm$ 1.88} & 52.94 \scriptsize{$\pm$ 1.09} & 44.66 \scriptsize{$\pm$ 2.06} & \textcolor{black}{56.53 \scriptsize{$\pm$ 2.36}} & \textcolor{black}{55.99 \scriptsize{$\pm$ 1.11}} & \textbf{57.06 \scriptsize{$\pm$ 0.77}} \\
CIFAR-100 & 22.92 \scriptsize{$\pm$ 0.95} & 23.46 \scriptsize{$\pm$ 1.05} & 23.35 \scriptsize{$\pm$ 1.12} & 22.13 \scriptsize{$\pm$ 1.34} & 22.55 \scriptsize{$\pm$ 1.28} & 11.87 \scriptsize{$\pm$ 0.77} & \textcolor{black}{21.57 \scriptsize{$\pm$ 0.46}} & \textcolor{black}{19.91 \scriptsize{$\pm$ 1.00}} & \textbf{24.82 \scriptsize{$\pm$ 0.88}} \\
\bottomrule
\end{tabular}
\end{adjustbox}
\end{table*}
We present experimental results evaluating \texttt{\shortname} in heterogeneous FL scenarios.
We employ established FL benchmark datasets \cite{caldas2018leaf, li2020federated}\textcolor{black}{; additional experiments on a synthetic dataset specifically designed for a controlled federated linear regression analysis, made publicly available to support result validation and reproducibility, are reported in Appendix~\ref{app:linreg-app}.}
\textcolor{black}{A detailed description of dataset settings, parameters used, and model architectures is provided in Appendix~\ref{app:exp}. }
In our analysis, we compare \texttt{\shortname} against a diverse set of FL baselines.
From client selection methods, we include the uniform random policy of \texttt{FedAvg} \cite{mcmahan2017communication}, the exploitative \texttt{Power-of-Choice} strategy \cite{cho2022towards} that favors clients with higher local losses, and \texttt{ActiveFL} \cite{goetz2019active}, which balances exploration and exploitation through probabilistic selection.
From the broader FL literature, we also consider \texttt{FedProx} \cite{li2020federated}, a regularization-based approach designed for heterogeneous settings, and two clustering-based personalized methods, \texttt{IFCA} \cite{ghosh2020efficient} and \texttt{FeSEM} \cite{long2023multi}. \textcolor{black}{We additionally implement two recent client-selection policies: \texttt{FedCBS}~\cite{zhang2023fedcbs}, which promotes class-balanced participation, and \texttt{Oort}~\cite{lai2021oort}, which selects clients according to a utility score combining statistical informativeness and system efficiency; both are evaluated under the same experimental protocol as all other baselines.}
\textcolor{black}{Section~\ref{sec:classif} presents the comparison on classification tasks on real-world datasets; Section~\ref{sec:ablation} analyzes the sensitivity to the participation rate; finally, in Section~\ref{sim:extra kernel} some extra simulations are presented to further validate our choice for the proximity matrix.}
\subsection{FL Classification Benchmarks}\label{sec:classif}
We evaluate \texttt{\shortname} on standard FL classification benchmarks.
We consider MNIST \cite{lecun1998mnist}, CIFAR-10, and CIFAR-100 \cite{krizhevsky2009learning}, each partitioned across clients using a Dirichlet distribution with concentration parameter $\alpha = 0.1$ \cite{caldas2018leaf}, which introduces both label and quantity heterogeneity.
Each dataset is trained with $P = 10$ clients sampled per round and $S = 10$ local epochs. \textcolor{black}{We set $T = 100$ communication rounds for CIFAR-10 and CIFAR-100, and $T = 200$ for MNIST. The training horizon is chosen per dataset to correspond to a comparable computational budget: MNIST is trained with a lightweight multilayer perceptron on low-dimensional inputs, so that its per-round cost is much smaller than that of the convolutional networks used for CIFAR-10/100, and $200$ MNIST rounds require approximately the same wall-clock time as $100$ CIFAR rounds. This choice does not affect the fairness of the comparison, since within each dataset all methods are evaluated under identical settings ($T$, $P$, $S$, model architecture, and data partition).}
Model architectures and parameters are detailed in Appendix \ref{app:exp}.
Table~\ref{tab:class} reports average test accuracy~\cite{hossin2015review}, averaged across clients (the larger the better).
\textcolor{black}{Our results suggest that \texttt{\shortname} achieves the best or highly competitive performance across all benchmarks.}
In particular, it delivers notable gains over random client sampling (\texttt{FedAvg}) and over the loss-aware \texttt{Power-of-Choice}, which can fail to generalize well in heterogeneous regimes.
Regularization with \texttt{FedProx} and clustering-based methods such as \texttt{IFCA} and \texttt{FeSEM} yield competitive results, but remain below \texttt{\shortname}.
This confirms the effectiveness of our selection strategy in both simple and highly heterogeneous datasets.
\textcolor{black}{Results in Table~\ref{tab:class} are reported as mean $\pm$ twice the standard deviation across independent runs; the bands thus describe the run-to-run dispersion of each method. We observe that the margins naturally shrink on MNIST, an almost saturated benchmark on which all heterogeneity-aware baselines perform similarly, and the advantage of \shortname\ over the strongest baseline is small in absolute terms; the gains instead widen on the harder CIFAR-10 and CIFAR-100 tasks, where statistical heterogeneity has a stronger impact. Furthermore, beyond the average accuracy, \shortname\ consistently exhibits the smallest dispersion among all competitive methods in every benchmark.}
\textcolor{black}{Concerning the newly added selection baselines, \texttt{Oort} remains below \shortname\ on all benchmarks and exhibits markedly larger run-to-run dispersion, reflecting its greedy loss-driven selection. \texttt{FedCBS} attains performance comparable to \shortname\ on the $10$-class benchmarks, but falls clearly behind on CIFAR-100: its class-rebalancing objective structurally degrades whenever the number of classes exceeds the participation budget ($P < B$), a regime excluded by its own design guidelines~\cite{zhang2023fedcbs} yet common in realistic deployments. Moreover, \texttt{FedCBS} requires the server to know the clients' label distributions---or trusted-hardware/homomorphic-encryption infrastructure to compute their pairwise products privately---and is defined only for classification tasks. \shortname\ requires neither: it operates solely on the model updates that the server already receives in standard FL, applies to arbitrary learning tasks (cf.\ the regression experiments in Appendix~\ref{app:linreg-app}), and adapts its selection to the evolving training dynamics rather than to a static dataset property, making it more readily deployable in privacy-sensitive IoT scenarios.}
\subsection{Sensitivity to the Participation Rate}\label{sec:ablation}
\begin{figure}[t]
\centering
\includegraphics[width=.5\textwidth]{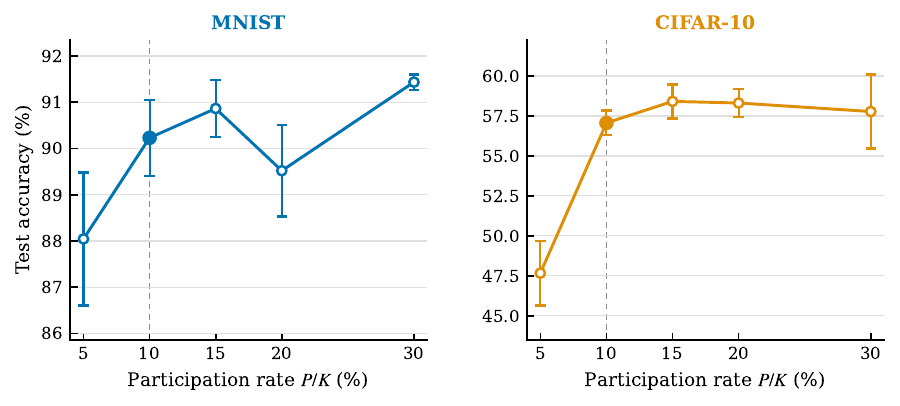}
\caption{\textcolor{black}{Final test accuracy of \shortname\ on MNIST (left) and CIFAR-10 (right) as a function of the participation rate $P/K$, varied from $5\%$ to $30\%$ ($K=100$). Points report mean $\pm 2$ standard deviations across independent runs; the filled marker denotes the $P=10$ operating point used in the main experiments (Table~\ref{tab:class}). Accuracy saturates beyond the $10\%$ rate, which attains near-peak performance at the lowest communication cost.}}
\label{fig:ablation_P}
\end{figure}

\textcolor{black}{Recall that in \shortname\ the number of participating clients $P$ coincides with the number of coalitions, so that $P/K$ is the participation rate and $K/P$ the average coalition size. In the main experiments we fix $P=10$, i.e., a $10\%$ participation rate, which is the common setup in FL benchmarks~\cite{caldas2018leaf}. Figure~\ref{fig:ablation_P} reports the final test accuracy of \shortname\ on MNIST and CIFAR-10 as the participation rate varies from $5\%$ to $30\%$. Accuracy degrades markedly below the $10\%$ rate, where too few coalitions are available to cover the heterogeneous client population, and saturates above it: raising the rate to $30\%$ yields no significant gain while tripling the per-round communication cost. Larger rates also shrink the average coalition size, making the within-coalition selection progressively closer to uniform sampling and exposing individual clients as quasi-deterministic representatives. The $10\%$ operating point thus achieves near-peak accuracy at the lowest communication budget, consistent with the limited-participation regime that \shortname\ is designed for.}
\subsection{Proximity matrix validation} \label{sim:extra kernel}
Let us now validate our choice for the proximity matrix through an ablation study to analyze our framework's robustness. We implement our algorithm with other similarity functions used in coalition formation \cite{Scholkopf2002}:
\begin{itemize}[leftmargin=*]
    \item \textbf{Cosine Similarity}: $W_{kj} = {\theta_k(t)^\top \theta_j(t)}/({||\theta_k(t)||_2 ||\theta_j(t)||_2})$
    \item \textbf{Laplacian Kernel}: $W_{kj} = \exp(-\gamma ||\theta_k(t) - \theta_j(t)||_1)$
    \item \textbf{Sigmoid Kernel}: $W_{kj} = \tanh(\gamma\theta_k(t)^\top \theta_j(t) + c)$
\end{itemize}
 While our theoretically-motivated choice yields the best performance ($90.23\%$ on MNIST, $57.06\%$ on CIFAR-10), our framework continues to substantially outperform the baselines even with the alternative similarity metrics as shown in Table~\ref{tab:proxy_metrics}.
\begin{table}
\centering
\caption{Test accuracy across proxy metrics. Best per dataset in bold.}
\label{tab:proxy_metrics}
\setlength{\tabcolsep}{4pt}
\scriptsize
\begin{tabular}{lcccc}
\toprule
 & \textbf{RBF (ours)} & \textbf{Cosine Sim.} & \textbf{Laplacian Ker.} & \textbf{Sigmoidal Ker.} \\
\midrule
\textbf{MNIST}    & \textbf{90.23 $\pm$ 0.82} & 88.72 $\pm$ 0.97 & 88.28 $\pm$ 0.65 & 88.92 $\pm$ 0.85 \\
\textbf{CIFAR-10} & \textbf{57.06 $\pm$ 0.77} & 55.32 $\pm$ 0.78 & 49.85 $\pm$ 0.71 & 50.25 $\pm$ 1.12 \\
\bottomrule
\end{tabular}
\end{table}
\section{Conclusion}\label{sec:conclusion}
We introduced \shortname, a novel framework addressing heterogeneous FL by drawing foundational insights from opinion dynamics models. The core of \shortname lies in a client selection strategy that biases sampling towards clients whose local models contribute most significantly to maximizing the variance reduction of the global model update. Distinct from conventional sampling methods, \shortname  exploits concepts from coalition formation within opinion dynamics to identify clusters of clients exhibiting similar model characteristics, i.e., the ``opinions.'' A representative client is then sampled from each identified cluster. This approach is supported by theoretical results that provide a foundation for our algorithm. Empirical evaluations on heterogeneous datasets, including a synthetic linear regression task specifically designed to highlight heterogeneity challenges and more complex benchmarks, demonstrate that our algorithm \textcolor{black}{outperforms or matches state-of-the-art} FL client selection algorithms.
\textcolor{black}{The analysis done opens several directions for further methodological and theoretical developments. First, a further extension would be to jointly optimize the number of clusters and clients, rather than fixing it a priori, by incorporating more advanced clustering strategies and explicitly accounting for the system’s communication and computational constraints, as well as the composition with device-level constraints, such as hardware heterogeneity, by restricting each coalition to its resource-feasible members before sampling}
Although our method effectively leverages pairwise individual model correlation, extending this framework to hypergraphs would model more complex and hierarchical dependencies between coalitions. Furthermore, our current analysis assumes stationarity of model distributions near convergence; future work could incorporate gradient flow dynamics to fully characterize the transient evolution of client models and coalition formation. Finally, there is significant potential to bridge the gap between our heuristic Boltzmann exploration and a rigorous statistical mechanics analysis of system stability\textcolor{black}{, including an analytical characterization of the gradient bias and variance induced by the proposed selection policy}.
\bibliography{references}
\bibliographystyle{ieeetr}
\setcounter{lemma}{0}
\renewcommand{\thelemma}{A\arabic{lemma}}
\setcounter{assumption}{0}
\renewcommand{\theassumption}{A\arabic{assumption}}
\setcounter{proposition}{0}
\renewcommand{\theproposition}{A\arabic{proposition}}
\setcounter{remark}{0}
\renewcommand{\theremark}{A\arabic{remark}}
\setcounter{equation}{0}
\renewcommand{\theequation}{A\arabic{equation}}
\newpage

\appendix
\subsection{Proof of Theorem~\ref{th:conv}}\label{app:th-conv}
We focus on a single step of the stochastic gradient descent, having $\theta_{gl}(t+1) = \theta_{gl}(t) - \eta g(t)$, where $g(t)$ is the federated gradient produced by \shortname. Due to Assumption~\ref{a:1}, if the loss is $L$-smooth, we can apply the Descent Lemma from~\cite[Chapter~2, Theorem~2.1.5]{nesterov}, bounding
\begin{equation}\label{eq:bound1}
\mathcal{L}(\theta_{gl}(t+1)) \leq \mathcal{L}(\theta_{gl}(t))-\eta  \nabla\mathcal{L}(\theta_{gl}(t))^\top g(t) + \tfrac{L\eta^2  \|g(t)\|^2}{2}.
\end{equation}
By taking the expectation up to the $t$-th round
, denoted by $\mathbb{E}_t$, the bound in \eqref{eq:bound1} becomes
\begin{align}
 \mathbb{E}_t[\mathcal{L}(\theta_{gl}(t+1))] \leq &\mathcal{L}(\theta_{gl}(t))-\eta  \nabla\mathcal{L}(\theta_{gl}(t))^\top \mathbb{E}_t[g(t)] \nonumber\\&+ \tfrac{L\eta^2\mathbb{E}_t[\|g(t)\|^2]}{2}.\label{eq:bound2}\end{align}
Now, we derive an explicit bounds for the terms in~\eqref{eq:bound2}.
We can assume that after $T_0$ we are in a thermalized regime, where the spectral clustering has converged, and the clusters are stationary, which is a standard assumption in FL~\cite{kim24clustered}; namely, $\forall t > T_0, \, \mathcal{C}_p(t) = \mathcal{C}_p^*$. Hence, also the Boltzmann measure $\pi(t)$ is stationary. Let us recall that we defined $\mathbb{E}_t[g(t)] = G(\theta_{gl}(t))$. {\color{black}Based on Assumptions~\ref{a:1}--\ref{a:2}}, $\nabla\mathcal{L}(\theta_{gl}(t))^\top \mathbb{E}_t[g(t)] =  \nabla\mathcal{L}(\theta_{gl}(t))^\top G(\theta_{gl}(t)) \geq c\|\nabla \mathcal{L}(\theta_{gl}(t))\|^2$ and {\color{black}$\mathbb{E}_t[\|g(t)\|^2]\leq M$.
Hence,} we re-write \eqref{eq:bound2} as
\begin{equation}\label{eq:bound5}\mathbb{E}_t[\mathcal{L}(\theta_{gl}(t+1))] \leq \mathcal{L}(\theta_{gl}(t))-\eta c\|\nabla \mathcal{L}(\theta_{gl}(t))\|^2 + \tfrac{LM\eta^2}{2}.\end{equation}
By taking the expectation of \eqref{eq:bound5} over the whole iterations, denoted by $\mathbb{E}$, we get
\begin{equation}\label{eq:bound6}\mathbb{E}[\mathcal{L}(\theta_{gl}(t+1))] \leq \mathbb{E}[\mathcal{L}(\theta_{gl}(t))]-\eta c\mathbb{E}[\|\nabla \mathcal{L}(\theta_{gl}(t))\|^2] +  \tfrac{LM\eta^2}{2}.\end{equation}
By summing \eqref{eq:bound6} over times, from $T_0$ to $T-1$, and rearranging the terms in the, we get
\begin{align}
\eta c \sum\nolimits_{t = T_0}^{T-1}\mathbb{E}[\|\nabla \mathcal{L}(\theta_{gl}(t))\|^2] \leq(T- T_0) \tfrac{LM\eta^2}{2}\nonumber\\ \qquad+\sum\nolimits_{t = T_0}^{T-1} (\mathbb{E}[\mathcal{L}(\theta_{gl}(t))] - \mathbb{E}[\mathcal{L}(\theta_{gl}(t+1))]).\label{eq:bound7}
\end{align}
Expanding the telescoping sum term on the right-hand-side of \eqref{eq:bound7}, it reduces to $\mathbb{E}[\mathcal{L}(\theta_{gl}(T_0))] - \mathbb{E}[\mathcal{L}(\theta_{gl}(T))]$. Since the loss is uniformly bounded from below (Assumption~\ref{a:1}), $\mathbb{E}[\mathcal{L}(\theta_{gl}(T))] \geq \mathcal{L}^*$ and the expression in \eqref{eq:bound7} simplifies to
\begin{equation}\label{eq:bound8}
\eta c \sum_{t = T_0}^{T-1}\mathbb{E}[\|\nabla \mathcal{L}(\theta_{gl}(t))\|^2] \leq \mathbb{E}[\mathcal{L}(\theta_{gl}(T_0))] - \mathcal{L}^* + (T- T_0) \tfrac{LM\eta^2}{2}.
\end{equation}
Dividing both terms by $\eta c(T - T_0)$ the \eqref{eq:bound8} reads
\begin{equation}\begin{array}{l}\tfrac{1}{T - T_0} \sum\nolimits_{t = T_0}^{T-1}\mathbb{E}[\|\nabla \mathcal{L}(\theta_{gl}(t))\|^2] \leq \\ \qquad\qquad\tfrac{\mathbb{E}[\mathcal{L}(\theta_{gl}(T_0))]}{\eta c(T-T_0)} - \tfrac{\mathcal{L}^*}{\eta c(T-T_0)} + \tfrac{LM\eta }{2c}.\label{eq:bound9}\end{array} \end{equation}
Finally, we consider the limit $T \to \infty$. Being $T_0$ finite, then it is negligible with respect to $T$. Hence, by adopting a change of variable and taking the limit with respect to $T$, \eqref{eq:bound9} reads
\begin{equation}\limsup_{T \to \infty} \tfrac{1}{T} \sum\nolimits_{t = 0}^{T-1} \mathbb{E}[\|\nabla \mathcal{L}(\theta_{gl}(t))\|^2] \leq \tfrac{LM\eta}{2c},\end{equation}
which yields the claim. \qed
\subsection{Implementation Details and Further Experiments} \label{app:exp}
\subsubsection{Datasets and Implementation Details}

We evaluate \shortname on various datasets. For a more straightforward scenario, we consider a one-dimensional linear regression under both IID and non-IID distributions, with and without intercept, i.e., $D = 1$ and $D = 2$ respectively. Then, we evaluate \shortname using several classification benchmark datasets, widely used in FL \cite{li2020federated, caldas2018leaf}: the Synthetic dataset \cite{li2020federated, cho2022towards}, Federated MNIST \cite{lecun1998mnist}, and CIFAR-10 \cite{krizhevsky2009learning}. To simulate a heterogeneous environment, we employ the Synthetic(1,1) setting, while the other datasets are partitioned using a Dirichlet distribution with $\alpha = 0.1$, by implementing a sampler according to \cite{caldas2018leaf}. For MNIST, a multilayer perceptron was utilized, comprising two layers with 200 hidden neurons and ReLU activations. The CIFAR-10 and CIFAR-100 experiments involved a convolutional neural network, which consisted of two convolutional and max-pooling layers, followed by two fully connected layers, for generating class predictions.
In all experiments, the dataset is partitioned across $K = 100$ clients, followed by dividing the datasets into training and testing sets to assess model performance. The models have been trained with Stochastic Gradient Descent with $S = 10$ local epochs, a batch size of $100$, and a learning rate of $\eta = 0.01$. We set the number of communication rounds to $ T = 200$ for  MNIST and $ T = 100$ for the other datasets. For each round, $P = 10$ clients are sampled to participate in the training.
We employed $\gamma_t = 1/t$ in the training of \shortname to ensure the convergence of Robbins-Monro estimators applied to the covariance \cite{robbins1951stochastic}. In the case of \texttt{Power-of-choice} \cite{cho2022towards}, we obtained $ d = 2P$ samples and subsequently selected the $P$ that exhibited the highest test loss on the global model. For \texttt{Active FL} \cite{goetz2019active}, we set $\alpha_1 = 0.8$ (which is equivalent to $d = 2P$), and similarly utilized the temperature parameter $\alpha_2 = 1$ as our temperature parameter, $\alpha_3 = 0$ to achieve comparability with \texttt{Power-of-choice}.
Synthetic experiments were executed locally on an Apple M1 processor, while MNIST and CIFAR experiments utilized an \texttt{RTX8000 NVIDIA} GPUs. Code is available at \url{https://github.com/alelicciardi99/fedcvr_bolt}.
\subsubsection{Practical Implementation}
Practically, \shortname clusters client according to the observed model $\theta_k(t)$. {To further stabilize the initial estimation phase, } we implement a uniformly random sampling strategy during the initial 30 rounds, enabling the observation of local adaptations of the global model across various sampled clients. The selection of 30 rounds offers a balanced approach between exploring client heterogeneity and subsequently leveraging the clustering structure. \textcolor{black}{This value admits a simple probabilistic justification: under uniform sampling, each client is selected with probability $P/K$ at every round, independently across rounds, so the probability that a given client is observed at least once within $t$ rounds is $1-(1-P/K)^t$~\cite{grimmett2001probability}. With $P=10$ and $K=100$, choosing $t=30$ yields $1-(1-0.1)^{30}\approx 0.96$, i.e., more than $95\%$ of the clients are expected to be observed at least once during the warm-up, ensuring an almost complete coverage of the population before the variance-reduction policy is activated.}
This initial phase is dedicated to observing a diverse set of client behaviors, which allows the Robbins-Monro estimator to build a more reliable and stable baseline covariance structure before our variance-reduction sampling policy becomes active \cite{grimmett2001probability}.
Similar to numerous FL methods based on local updates~\cite{ghosh2020efficient} executing operations on the full model becomes increasingly burdensome as the model size increases. Furthermore, from a statistical standpoint, the early feature extraction layers (such as convolutional or embedding layers) frequently contribute less discriminative information for clustering and may introduce superfluous noise along with computational overhead. Hence, when managing larger models, we confine the covariance computation to the parameters of the final fully connected layer. If this layer remains excessively large, we further diminish dimensionality by randomly sampling a subset of its weights. Specifically, for the MNIST dataset, we utilize the entire final layer; whereas for CIFAR-10, where the final layer has higher dimensionality, we sample $D=300$ weights. This approach achieves an advantageous compromise between computational efficiency and representational adequacy.
\begin{table*}[t]
\centering
\caption{\small Comparison between \shortname and FL selection baselines on a synthetic regression dataset. }
\begin{adjustbox}{width = \textwidth}
\begin{tabular}{l|l|cccc}
\toprule
\textbf{Regression Model} & \textbf{Heterogeneity} & \texttt{FedAvg} & \texttt{Power-of-Choice} & \texttt{AFL} & \cellcolor{gray!10}\texttt{FedCVR-Bolt} \\
\midrule
\multirow{2}{*}{$y = \theta_1 x$}
  & IID      & 6.0171 \scriptsize{$\pm$ 2.4884}  &  6.3587 \scriptsize{$\pm$ 2.0280} & 6.1656 \scriptsize{$\pm$ 2.5821} & \cellcolor{gray!10}\textbf{6.0136\scriptsize{$\pm$ 2.5170}} \\
  & Non-IID  &57.1336 \scriptsize{$\pm$ 24.1335} &  56.9701 \scriptsize{$\pm$ 24.7835} &56.8328 \scriptsize{$\pm $ 25.7835}  & \cellcolor{gray!10}\textbf{54.7102 \scriptsize{$\pm$ 24.8973}} \\
\midrule
\multirow{2}{*}{$y = \theta_0 + \theta_1 x$}
  & IID      &0.2142\scriptsize{$\pm$0.3520}   &  0.2143 \scriptsize{$\pm$ 0.3533}&  0.2134 \scriptsize{$\pm$ 0.3503} & \cellcolor{gray!10}\textbf{0.2085 \scriptsize{$\pm$ 0.3386}} \\
  & Non-IID  & 1.5875 \scriptsize{$\pm $ 0.7291}   &  1.6742 \scriptsize{$\pm $ 0.6820} & 1.7023 \scriptsize{$\pm$ 0.7642}& \cellcolor{gray!10}\textbf{1.5127 \scriptsize{$\pm$ 0.7336}} \\
\bottomrule
\end{tabular}
\end{adjustbox}
\label{tab:linreg}
\end{table*}
\subsubsection{On the Boltzmann temperature}\label{app:temperature}
\textcolor{black}{We remark that the Boltzmann-like sampling measure $\pi_p(k;t)$ does not involve a free temperature parameter. Interpreting the variance-reduction score $v_k$ as the energy of the associated Gibbs measure, the sampling law is the maximum-entropy distribution consistent with these scores, for which the inverse temperature is fixed to its canonical value $\beta=1$: any rescaling of the temperature can be equivalently absorbed into the normalization of $v_k$, so that it does not constitute an independent degree of freedom. Its two extremes are nonetheless informative, as $\beta\to\infty$ recovers a greedy (argmax) selection of the most informative client, while $\beta\to 0$ recovers uniform sampling within each coalition; the canonical choice $\beta=1$ balances these regimes without introducing an additional hyperparameter to tune. For this reason, the sensitivity analysis of Section~\ref{sec:ablation} focuses on the participation rate $P$, which is the only genuine hyperparameter of the selection rule.}
\subsubsection{Federated Linear Regression}\label{app:linreg-app}

\textcolor{black}{As a further experiment, we evaluate \shortname\ on a controlled federated linear regression task, which also illustrates the applicability of the method beyond classification.} To simulate client heterogeneity, we generate synthetic datasets across $K=100$ clients, where each client $k$ is assigned to one of $J$ latent clusters. Local inputs $x_k^i \in \mathbb{R}^D$ are sampled from a cluster-specific Gaussian $\mathcal{D}(\theta_{x, j_k}, \sigma_{x,j_k}^2 I_D)$, and labels are computed as $y_k^i = (\theta_k^i)^\top x_k^i$ using latent parameters $\theta_k^i \sim \mathcal{D}(\bar{\theta}_{j_k}, \sigma_{\theta,j_k}^2 I_D)$. We consider both IID ($J=1$) and non-IID ($J=2$) settings, handling intercepts by concatenating a bias component to the input. With training set to $T=100$ rounds, selecting $P=10$ clients per round with $S=10$ local SGD steps, Table~\ref{tab:linreg} shows that \texttt{\shortname} maintains competitive performance in IID scenarios and consistently outperforms baselines in non-IID settings, reducing test MSE by up to $3.0\%$ over uniform sampling.
\end{document}